\documentclass{article} 
\usepackage{iclr2025_delta,times}

\usepackage{microtype}
\usepackage{graphicx}
\usepackage{booktabs} 

\usepackage{xcolor}

\usepackage{hyperref}

\usepackage{multirow}

\usepackage{amsmath}
\usepackage{amssymb}
\usepackage{mathtools}
\usepackage{amsthm}
\usepackage{mathrsfs}

\usepackage[capitalize,noabbrev]{cleveref}

\usepackage[textsize=tiny]{todonotes}

\usepackage{amsmath,amsfonts,bm}

\def\eqref#1{equation~\ref{#1}}

\def\1{\bm{1}}

\DeclareMathAlphabet{\mathsfit}{\encodingdefault}{\sfdefault}{m}{sl}
\SetMathAlphabet{\mathsfit}{bold}{\encodingdefault}{\sfdefault}{bx}{n}

\newcommand{\E}{\mathbb{E}}

\DeclareMathOperator*{\argmin}{arg\,min}

\usepackage{hyperref}
\usepackage{url}

\usepackage{amsmath,amsthm,amssymb}
\usepackage{dsfont}
\usepackage{wrapfig}
\usepackage[noabbrev,capitalize]{cleveref}

\usepackage{bbm}

\usepackage{booktabs}

\usepackage{bm}
\usepackage{paralist}
\usepackage{enumerate}
\usepackage[inline]{enumitem}

\usepackage{mathtools}
\usepackage{listings}
\usepackage{xcolor}
\definecolor{codegreen}{rgb}{0,0.6,0}
\definecolor{codegray}{rgb}{0.5,0.5,0.5}
\definecolor{codepurple}{rgb}{0.58,0,0.82}
\definecolor{backcolour}{rgb}{0.95,0.95,0.92}
\lstdefinestyle{mystyle}{
    backgroundcolor=\color{backcolour},   
    commentstyle=\color{codegreen},
    keywordstyle=\color{magenta},
    numberstyle=\tiny\color{codegray},
    stringstyle=\color{codepurple},
    breakatwhitespace=false,         
    breaklines=true,                 
    captionpos=b,                    
    keepspaces=true,                 
    numbers=left,                    
    numbersep=5pt,                  
    showspaces=false,                
    showstringspaces=false,
    showtabs=false,                  
    tabsize=2
}
\usepackage{graphicx}

\usepackage{ifthen} %
\usepackage{soul}
\usepackage{subcaption} 
\usepackage{booktabs}
\usepackage{adjustbox} 
\usepackage{etoolbox}  
\usepackage{accents}
\usepackage{apptools}
\usepackage{float}
\usepackage{scalerel,stackengine}
\usepackage{mathtools}
\usepackage[most]{tcolorbox}
\usepackage{cleveref}
\usepackage{caption}
\usepackage{comment}
\usepackage{tikz}
\usetikzlibrary{intersections,arrows, arrows.meta, decorations.markings, matrix, calc, bayesnet, automata, chains, backgrounds, positioning, fit, shapes}
\usepackage{soul}
\newtheorem{theorem}{Theorem}

\newtheorem*{remark}{Remark}
\usepackage[capitalize]{cleveref}
\usepackage{tcolorbox}
\usepackage{adjustbox}
\usepackage{multicol}
\usepackage{multirow, booktabs}
\usepackage{longtable}
\usepackage{array}
\usepackage{tabularx}
\usepackage{calc}
\usepackage{makecell}
\usepackage{bm}
\usepackage{colortbl}
\usepackage{bigints}
\usepackage{algorithm}
\usepackage{algorithmic}

\usepackage{newfloat}
\usepackage{listings}

\theoremstyle{plain}
\newtheorem{proposition}[theorem]{Proposition}

\theoremstyle{definition}

\theoremstyle{remark}

\AfterEndEnvironment{rem}{\noindent\ignorespaces}

\def\model(#1,#2){\ensuremath{%
v_\theta(#1,\,#2)
}}
\def\phicexpow^#1#2#3{
    \phicexusualcommon{\phi^{#1}}{#2}{#3}}
\def\phicexusual#1#2{
    \phicexusualcommon\phi{#1}{#2}}
\def\phicexusualcommon#1#2#3{
    #1_{#2,#3}\@ifnextchar(\negmathsp{}}
\def\phicex{\@ifnextchar^\phicexpow\phicexusual}
\def\phic{\phi_{t,x_1}}

\def\psicexpow^#1#2#3{
    \psicexusualcommon{\psi^{#1}}{#2}{#3}}
\def\psicexusual#1#2{
    \psicexusualcommon\psi{#1}{#2}}
\def\psicexusualcommon#1#2#3{
    #1_{#2,#3}\@ifnextchar(\negmathsp{}}
\def\psicex{\@ifnextchar^\psicexpow\psicexusual}

\iclrfinalcopy

    \title{Learning Straight Flows by Learning Curved Interpolants}
    \author{
Shiv Shankar \\
University of Massachusetts
\\\texttt{sshankar@cs.umass.edu} \\
\And
Tomas Geffner \\
NVIDIA
\\\texttt{tgeffner@nvidia.com } \\
}

\begin{document}
    \maketitle
    
    \begin{abstract}
Flow matching models typically use linear interpolants to define the forward/noise addition process. This, together with the independent coupling between noise and target distributions, yields a vector field which is often non-straight. Such curved fields lead to a slow inference/generation process. In this work, we propose to learn flexible (potentially curved) interpolants in order to learn straight vector fields to enable faster generation. We formulate this via a multi-level optimization problem and propose an efficient approximate procedure to solve it. Our framework provides an end-to-end and simulation-free optimization procedure, which can be leveraged to learn straight line generative trajectories. 
\end{abstract}

%

\vspace{-0.5cm}
\section{Introduction}

In recent years, neural generative models have become remarkably successful in a variety of domains, including computer vision \citep{xing2023survey}, data generation \citep{karras2020analyzing}, robotics \citep{firoozi2023foundation}, and scientific applications \citep{guo2024diffusion}.
This success can be broadly attributed to the development of generative models based on learning vector fields  \citep{ho2020denoising,yang2023diffusion,lipman2022flow,albergo2022building}. 
Among these methods, flow matching \citep{lipman2022flow,albergo2022building} is a recent simulation-free method that has gained traction due to its robust performance. Flow matching is a version of Continuous
Flows (CNFs) \citep{chen2018neural} that uses deep neural networks to learn a velocity vector field that transports samples from a source distribution to the target
distribution. \citet{peluchetti2022nondenoising,lipman2022flow} show that a regression objective on conditional probability paths between source and target samples can be used to learn a model of the velocity field without an explicit target field.

Despite their success, flow-matching models (as well as other vector-field-based models) often suffer from slow sampling. Specifically, one has to simulate an ODE (or SDE) with a numerical solver to generate samples. Since these models learn a vector field that needs to be numerically integrated, the overall process can be slow and rife with discretization errors \citep{song2023consistency}. This is because the learnt vector fields often have curved trajectories, and therefore small discretization step-sizes are needed for accurate simulation. 

To alleviate this issue, many methods have been proposed to learn more easily integrable vector fields. One common approach tries to incorporate `straightness' to the field \citep{liu2022flow,liu2023instaflow,kornilov2024optimal,tong2023improving}, as straight paths allow for 1-step and discretization-free inference. While few proposals learn straight approximations to the true field \citep{yang2024consistency,zhang2024mutli,lee2023minimizing}, most of these share a common flaw. Most of these these methods use linear conditional paths and independent coupling during training. While easy to implement, the underlying vector field in these scenarios is fundamentally curved, and any `straight' version will be necessarily approximate.

In this work, we propose a new method to learn straight flows using flow matching by tuning interpolants. We propose explicitly training the conditional probability paths (or interpolants) used in flow matching to enforce straightness in the velocity field. While mathematically simple, the corresponding bi-level optimization problem is intractable. We address this by proposing an approximate approach to solve this optimization problem, which relies on an an analytic form for the target vector field of a Conditional Flow Matching objective \citep{lipman2022flow}, and enforcing straightness on the target field tuning the interpolant. Empirically, we observe that our method outperforms recent models on standard datasets.

\paragraph{Contributions} (a) We propose a new bi-level formulation to learn straight flows that explicitly forces straightness by learning interpolants, (b) We derive an analytic form of target vector field that allows training interpolants without differentiable optimization methods, (c) We present scalable parametric models for conditional paths that enable efficient training, (d) We show significant improvements for low-shot (i.e., low number of function evaluations) generation quality.

\begin{figure*}[t]
  \centering
  \begin{minipage}[b]{0.45\textwidth}
    \centering
    \includegraphics[width=0.95\textwidth,trim={0cm 0 18cm 0cm},clip]{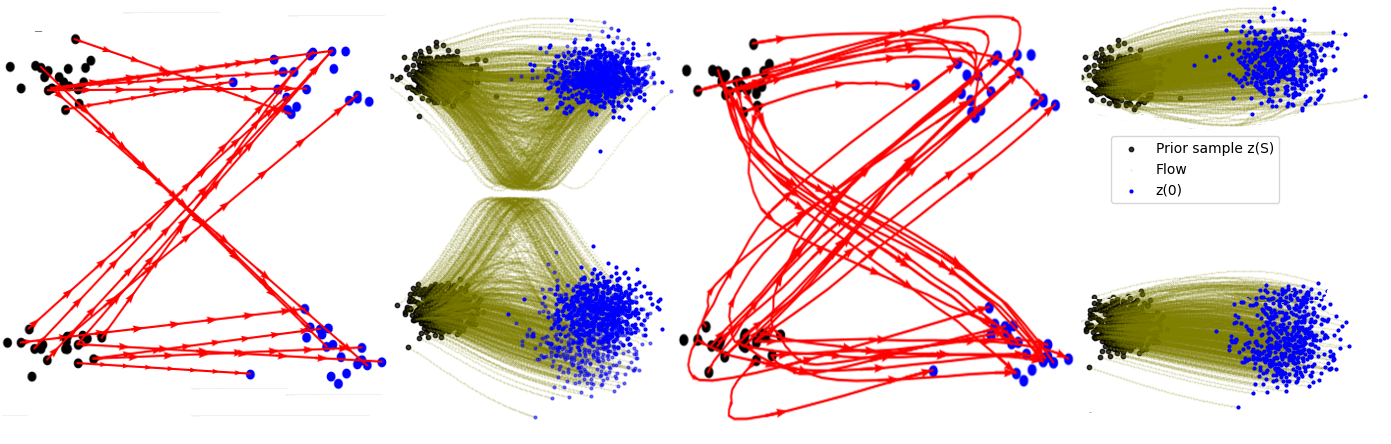} 
    \caption*{FM with Linear Interpolants}
    \label{fig:example1}
  \end{minipage}
  \hfill
  \begin{minipage}[b]{0.45\textwidth}
    \centering
    \includegraphics[width=0.96\textwidth,trim={17.2cm 0 0cm 0cm},clip]{figs/fig_example1.png} 
    \caption*{FM with Learned Interpolants (ours)}
    \label{fig:example2}
  \end{minipage}
  
  \caption{Illustration of training difference flow matching (FM) \citep{lipman2022flow} and our method on an illustrative 2 gaussian (source, black) to 2-gaussian (target, blue) problem. On the left we have standard FM with linear interpolants (red lines) and the resulting flow vector. On the right we show our approach. By allowing non-linear interpolants we are able to learn a flow velocity field with significantly less curvature. 
  \label{fig:first_fig}}
\end{figure*}


\section{Preliminaries}

Let  $p_0$ and $p_1$ be two distributions on $\mathbb{R}^d$. Usually, $p_1$ is our target distribution, only known through samples, and $p_0$ is the source distribution, usually chosen to be tractable (e.g., a Gaussian). Our goal is to generate samples from $p_1$. One way to generate these samples is by transporting the initial distribution $p_0$ via a vector field to the target $p_1$. Specifically, consider a vector field $u(t,x): [0,1]\times\mathbb{R}^d \xrightarrow{} \mathbb{R}^d$ such that the differential equation

\begin{equation}
\label{ODE-1}
\frac{d \gamma_x(t)}{dt} = u(t,\gamma_x(t)), \qquad
\gamma_x(0) \sim p_0
\end{equation}
produces sample from $p_1$. Under some topological constraints on the distributions, the existence of such fields is well known \citep{ambrosio2014continuity}. For a given starting sample $x$, we denote the solution of \cref{ODE-1} as $\gamma_x(t)$. This solution, also called a flow, is the trajectory of point $x$ as it evolves under $u$. We denote by $p_t$ the probability distribution obtained at time $t$ by moving the samples of $p_0$ via the field $u$. Note that for a given source and target distributions, multiple vector fields which satisfy the above equation exist.

\subsubsection*{Flow Matching}
Flow matching (FM) \citep{lipman2022flow} is a simulation-free method for learning one such vector field using a neural network $v_\theta(t,x)$. It does that by optimizing the FM objective
\begin{equation} \label{objective-1}
    \mathcal{L}_{FM} = E_{t,p_t} ||v_\theta(t,x_t)-u(t,x_t)||_2^2.
\end{equation}
Unfortunately, this objective is not tractable given only the source and target distributions, as the target vector field $u(t,x_t)$ is unknown. \citet{lipman2022flow} show that one can optimize the conditional flow matching objective instead

\begin{equation} \label{eq:Lcfm}
    \mathcal{L}_{CFM} = E_{t,q(z)}E_{p_t(x_t|z)} ||v_\theta(t,x_t)-u(t,x_t|z)||_2^2,
\end{equation}
where $u(t,x_t|x_1)$ is a conditional vector field and corresponding probability path $p_t(x_t|z)$. \citet{lipman2022flow} considered $z$ to be a sample from the target $x_1$. While \citet{lipman2022flow} considered a family linear Gaussian paths, more generalized variants of this problem have been proposed \citep{tong2023improving}.
Any suitable conditioning variable $z$ can be chosen if the objective remains tractable \citep{pooladian2023multisample,tong2023improving}.
A related problem is that of the Schrodinger Bridge \citep{de2021diffusion,wang2021deep}, which seeks the vector field whose probability law is close to that of the standard Brownian diffusion process.

\section{Learning Straight Flows by Tuning Interpolants}

In this section we describe our method to learn straight flows. We start with the standard CFM objective, but introduce parametric interpolants (parameterized by $\phi$), instead of the linear ones often used in flow matching. Once we do so, the learned flow model $v_\theta$ becomes dependent on the interpolant parameters $\phi$. We thus formulate the problem of learning straight flows as a bi-level optimization problem, where the interpolants are tuned to optimize the straightness of $v_\theta$ (\cref{sec:bilevel}). Next, we show how one can solve this bilevel optimization without using differentiable optimization methods. For this purpose, we derive an analytic expression for the flow field in terms of the parameters $\phi$ (\cref{sec:ref_fm}) together with a measure of the flow's straightness (\cref{sec:str}). Finally, we describe a specific family of interpolants which enables scalable computation with high-dimensional datasets (\cref{sec:scalability}).

\subsection{Bi-Level Formulation for Straight Flows} \label{sec:bilevel}
We consider the commonly used CFM objective with $z=(x_0,x_1)$ \citep{tong2023improving}. While CFM models often use simple linear interpolants, the consistency of the CFM approach only requires smooth enough conditional fields. Crucially the linear interpolant $x_t$ can be replaced by any other  interpolant (stochastic or otherwise), as long as the conditional $u$ (\cref{eq:Lcfm}) is modified to be the corresponding tangent vector. Specifically, choosing the interpolant to be $x_t = \phi_{t,x_1}(x_0)$ we can rewrite the CFM objective from \cref{eq:Lcfm} as 
\begin{equation} \label{objective-3}
    E_{t,p_0(x_0),p_1(x_1)}||v_\theta(t,x_t)- \partial_t \phi_{t,x_1}(x_0)||_2^2,
\end{equation}
which recovers standard CFM for $\phi := tx_1 + (1-t)x_0$.

We propose learning straight flows by choosing $\phi$ such that the resulting vector field $v$ is as straight as possible. We formulate this as a bi-level optimization problem
\begin{align} \label{objective-bilevel}
    &\qquad \qquad \qquad  \qquad \min_\phi |v_{\theta^*}(.;\phi)|_\text{Straight} \qquad  \text{s.t.} \\ 
    &v_{\theta^*} = \argmin_\theta E_{t,p_0(x_0),p_1(x_1)}||v_\theta(t,x_t; \phi)- \partial_t \phi_{t,x_1}(x_0)||_2^2, \nonumber
\end{align}
where $|\cdot|_\text{Straight}$ is a measure of straightness of the vector field. As an example, previous work by \citet{pooladian2023multisample,liu2022flow} suggest measuring curvature using

\[
|v|_\text{Straight} = \mathbb{E}_{\substack{t \sim U(0,1) \\ x_0 \sim p_0}} \left[ ||v_t(\gamma_{x_0}(t))||^2 - ||\gamma_{x_0}(1) - x_0||^2 \right],
\]
where $\gamma$ is the flow, that is, the solution to Equation 1.
\begin{remark}
Note that $v_\theta$ is a function of $\theta$, but also has an implicit dependence on $\phi$. Due to the inner optimization, the optimal parameter $\theta^*$ depends on $\phi$. We made this dependence explicit in \cref{objective-bilevel}. 
\end{remark}
While mathematically straightforward, solving the optimization problem from \cref{objective-bilevel} is computationally intensive.
This is a bi-level problem, where each inner objective requires solving a FM objective. Our insight is that one is not required to solve the FM objective to get $v_{\theta^*}$, and can instead leverage the optimal vector field $v_{\theta*}$ in a different manner directly in terms of the interpolant functions $\phi$.

\subsection{Reformulating Flow Matching} \label{sec:ref_fm}

This section shows how to connect $v_\theta$ to the interpolant function $\phi$ without solving the inner optimization iteratively, by presenting an analytic form for the optimal $v_\theta^*$ (\cref{eq:opt}) in terms of $\phi$. We will then use this result in \cref{sec:str} and \cref{sec:scalability} to derive our final optimization algorithm.

\begin{proposition} \label{prop:v_expr}
Let $\mathcal{J}$ be the determinant of the Jacobian of the interpolant's inverse, i.e., $\left| \frac{d \phic^{-1}(x)}{dx}\right|_\Delta$. The optimal velocity field for an interpolant $\phi$ is given by
\begin{align}
v_\phi^*(t,x_t) =  \int \dfrac{ \partial_t \phic(\phic^{-1}(x_t)) p_0(\phic^{-1}(x_t)) |\mathcal{J}| p_1(x_1) dx_1}{ \int p_0(\phic^{-1}(x_t)) |\mathcal{J}|  p_1(x_1)dx_1 }.
\label{eq:opt}
\end{align}
\end{proposition}

Specific simple parameterisations, like paths of the form $x_0 = \frac{x_t - g(t) x_1}{1- h(t)}$ with the coefficients $g,h$ being non-linear in $t$, yield efficient versions of \cref{eq:opt}. For such paths, \cref{eq:opt} simplifies as the Jacobians do not depend on $x_1$. However, such parameterisations, while simple to use, may not be powerful enough for complex datasets.

While exact, \cref{eq:opt} is intractable, as it involves complex integrals. In practice, our final algorithm will leverage empirical estimates for this quantity using samples from $p(x_1)$, and flexible interpolants $\phi$ parameterized by neural networks, as explained in \cref{sec:scalability}.

\subsubsection{Derivation of \cref{eq:opt}} \label{sec:proof_prop}

A key challenge in analyzing the FM objective is that both terms involved in the $L2$ loss (\cref{objective-3}) depend on $x_0$ and $x_1$. This can be addressed by writing out the distribution over $x_t$ explicitly 
\begin{align} \label{objective-4}
    &E_{t,p_0(x_0),p_1(x_1)}||v_\theta(t,x_t)- \partial_t \phi_{t,x_1}(x_0)||_2^2 \\
    &= E_{t,p(x_0,x_1)}||v_\theta(t,x_t)- \partial_t \phi_{t,x_1}(x_0)||_2^2  \\
    &= E_{t,p_t(x_t),p(x_0,x_1|x_t,t)}||v_\theta(t,x_t)- \partial_t \phi_{t,x_1}(x_0)||_2^2,
\end{align}
where it can be observed that the global minimizer is
\begin{align}
v^*(t,x_t) = \E_{p(x_0, x_1 | x_t, t)}[ \partial_t \phi_t(x_0, x_1)]. \label{eq:vopt}
\end{align}
This expression can also be directly obtained from the generalized CFM objective \citep{pooladian2023multisample}, using $z=x_t$. 


While simple, \cref{eq:vopt} cannot be used in practice since we do not have access to the required distributions. For instance, $x_1$ is only available via samples, and the posterior $p(x_0, x_1 | x_t)$ is intractable. This can be addressed by some analysis and a few judicious choices conditional paths. First we write the conditional distribution as:
\begin{align}
p(x_0,x_1|x_t,t) &= \frac{p(x_0,x_1, x_t|t)}{p(x_t|t)} = \frac{p(x_0,x_1, x_t|t)}{ \int p(x_0,x_1,x_t|t) dx_0 dx_1}. \label{eq:cond_dist}
\end{align}
Specifically, using $x_t = \phic(x_0)$, \cref{eq:cond_dist} becomes
\begin{align}
p(x_0,x_1|x_t,t) &=  \frac{p(x_0,x_1, \phic(x_0)|t)}{ \int p(x_0,x_1,\phic(x_0)|t) dx_0 dx_1}.
\end{align}

Since the maps are deterministic, we get
\begin{equation}
p(x_0, x_1 \mid x_t, t) = \frac{\delta(\tilde{x}_t - x_t) \, p_0(x_0) \, p_1(x_1)}{p(x_t|t)},
\end{equation}
where $\tilde{x}_t = \phic(x_0)$ is an auxiliary variable introduced to distinguish the conditioning variable from the conditioned value, $\delta$ is the Dirac function \citep{dirac1981principles}, and $p(x_t)$ is the marginal likelihood (normalization constant), given by
\begin{align}
p(x_t | t) &= \int p(x_0, x_1, x_t| t) dx_0 dx_1 =  \int \delta(\tilde{x}_t - x_t) \, p_0(x_0) \, p_1(x_1) \, dx_0 \, dx_1. \label{eq:sub}
\end{align}
While the above integral can be estimated with infinite data, the expression is not directly practically useful, since the estimator will often just be zero when relying on empirical samples (due to the Dirac function).
However, a useful expression can be obtained noting that the Dirac delta function allows us to integrate over $x_0$ using the change in variables formula \citep{halperin1952introduction}, yielding
\begin{equation}
p(x_t| x_1, t) = p_0(\phic^{-1}(x_t)) \mathcal{J},
\end{equation}
where $\mathcal{J}$ is the determinant of the Jacobian of the interpolant's inverse, i.e., $\left| \frac{d \phic^{-1}(x)}{dx}\right|_\Delta$.
Substituting this in \cref{eq:sub} and using the definition of $v^{*}$ (the expectation of $\partial_t \phic$ under $p(x_0,x_1|x_t)$) yields \cref{eq:opt}.

\subsection{Straightening the Flow} \label{sec:str}
\Cref{eq:opt} yields powerful way to manipulate the target velocity field $v^*$ to enforce desired properties on the flow. When we want the flows to be straight, the vector field $v^*$ should be constant everywhere on a trajectory. This implies that a straight velocity field $u_{\text{str}}$ must satisfy 
\[
u_{\text{str}}(t, \gamma_{x_0}(t)) = u_{\text{str}}(s, \gamma_{x_0}(s)) = u_{\text{str}}(0, \gamma_{x_0}(0)) \,\, \forall s,t,
\]
where $\gamma_{x_0}$ is the flow generated by $u_{\text{str}}$ from point $x_0$. Since the vector field is constant, we can differentiate the above expression with respect to $t$ to get
\begin{align}
\partial_t u_{\text{str}} = - \nabla_x u_{\text{str}}  \cdot u_{\text{str}}
\label{eq:optvec}
\end{align}
for all points on the trajectory, where $\cdot$ refers to the matrix-vector product. Therefore, since we know the optimal field is given by $v^*$, we can learn straight flows by optimizing
\begin{align}
\phi^* = \arg\min_\phi || \partial_t v_\phi^* +  \nabla_x v_\phi^* \cdot v_\phi^*  ||^2.
\label{eq:regloss}
\end{align}

We note here that this criteria for straight flows has been observed before \citet{liu2022flow,yang2024consistency}.

While the above expression is correct, optimizing it is computationally intensive, as we need to differentiate through $v_\phi^*$, which in turns requires the Jacobian of $\phic^{-1}$. However, using a flexible neural network for $v_\theta$, under ideal training we expect $v_\theta \approx v_\phi^*$. Therefore, we propose to replace the derivatives of $v^*$ with derivatives of $v_\theta$, which yields
\begin{align}
 \min_{\theta, \phi} \E_t \E_{x_0,x_1} & || v_\theta(x_t,t) - \text{sg}(\partial_t \phi_{t,x_1}(x_0)) ||^2 \label{obj} + \lambda || \partial_t v_\theta(x'_t,t) +   \nabla_x v_\theta(x'_t,t) \cdot v^*(x'_t,t) ||^2. \ref{obj}
 \end{align}
Here ``$\text{sg}$'' refers to stop gradient and $\lambda >0$ is a hyper-parameter. We suppress the dependence of $v^*$ on $\phi$ for notational convenience.

\subsection{Scalable Computation} \label{sec:scalability}
While \cref{obj} is `analytical', in general, it has poor scaling in the dimensionality of the output, specifically due to the presence of the inverse-derivative as well as the Jacobian, both of which do not generally scale with the dimension of the data. Additionally, we need to ensure that the function $\phic$ remains invertible. Fortunately, researchers have developed models that are amenable to these considerations. Specifically, the literature on normalising flows and invertible models \citep{papamakarios2021normalizing} has proposed several families of expressive neural networks that support such operations. In this work we use GLOW/1x1 convolution model \citep{kingma2018glow} to parameterise the function $\phi$. This family of neural networks learns any linear transform's parameters in the ``PLU'' format. It initializes any matrix parameter $W$ of a linear layer, finds its PLU decomposition, fixes $P$ and optimizes the lower and upper diagonal matrices $L$ and $U$. Normalizations and activations are also chosen in a way that ensures easy inversion and memory efficiency.

\paragraph{Computing Expectations}
The expression for $v^*$ from \cref{eq:opt} requires computing expectations (or integrals) over the data distribution $p(x_1)$. Since $p_0$ is Gaussian (or a similar tractable distribution), the value of $p_0(\phic(x_t))$ can be computed exactly. Therefore, one can estimate both the numerator and the denominator as empirical expectations over the target distribution. The resulting estimator is consistent, with the bias going to zero as we aggregate more data samples. Additionally, this estimation process, which combines vector fields from multiple paths at the same time, often produces a lower variance estimate of the optimal $v^*$. In fact, \citet{xu2023stable} used this idea to reduce the variance of the gradient estimates used to train diffusion models, observing that it often led to more robust optimization, thanks to the variance reduction, despite the bias introduced.

\begin{algorithm*}[t!bh]
    \caption{Vector field model training algorithm}
    \label{alg:main}
	\begin{algorithmic}[1]
        \REQUIRE{Sampler for $p_0$ (usually Gaussian), Empirical samples from $p_1$, batch size $N$, averaging size $M (\geq N)$ (to estimate \cref{eq:opt});
            models~$v_\theta(x,t)$ and $\phi$
        }

        \WHILE{not converged}
        \STATE{Sample $N$ points $\{t^i\}_{i=1}^N$ from  $\mathcal U[0,1]$}
        \STATE{Sample $N$ pairs $\{x_0^i,x_1^i\}_{i=1}^n$ from $p(x_0, x_1) = p_0(x_0)p_1(x_1)$}
        \STATE{Sample $M-N$ points $\{\hat x_1^j\}_{i=N}^{M-N}$ from $p_1$ 
        }
        \STATE{Compute $N$ interpolants $x_t^i$ from $(t^i,x_0^i,x_1^i)$}
        \STATE{Estimate $v^*(x_t^i)$ (Eq.~\ref{eq:opt}) replacing integrals $\int f(x_t,x_1,t) p_1(x_1) dx_1$ by empirical estimates $\frac{1}{M} \sum_{j=1}^M f(x_t^i,x_1^j,t^i)$}
        \STATE{Calculate the empirical loss using \cref{obj}}
        \STATE{Update parameters~$\theta,\phi$ (e.g., using SGD or Adam \cite{kingma2014adam})}
        \ENDWHILE
	\end{algorithmic}
\end{algorithm*}

\section{Related Work}


Multisample FM \citep{pooladian2023multisample} proposes to generalize the independent coupling of the data distribution $p_1(x_1)$ and prior distribution $p_0(x_0)$ to the optimal transport coupling plan $\pi(x_0,x_1)$. Under the optimal transport plan, the learned trajectory of ODE are straight \citep{pooladian2023multisample}. However, this requires constructing the optimal transport plan for the data which is computationally prohibitive. Minibatch OT and Minibatch Sinkhorn coupling have been suggested to lower the cost of computing such couplings \citep{tong2023improving}. 

\citet{liu2022flow} suggested a rectified flow matching method which uses a pretrained FM to learn a straighter approximation. Based on this insight, other methods to learn straight flows have also been proposed \citep{liu2022flow,liu2023instaflow,lee2023minimizing}. These methods however are not simulation-free (requiring sampling during training) and often require iteratively distilling from flows, both of which are computationally intensive.

\paragraph{Non-Linear Interpolants} 
Rectified and distilled flow methods \citep{lee2023minimizing,liu2022flow} rely on the coupling given by a ``teacher'' (pre-trained) flow. These methods often rely on linear interpolants, and require multiple training stages, where each stage requires generating training pairs of noise and data samples by simulating the ODE with the vector field learnt in the previous stage. Our method, on the other hand, is simulation-free and uses the independent coupling, though with learned interpolants.
\citet{kapusniak2024metric} also proposed using non-linear interpolants. They learn an interpolant such that the conditional paths stay close to a manifold, but do not directly optimize any property of $v^*$. \citet{bartosh2024neural} proposed learning the forward process in diffusion models, akin to learning the interpolant in flow matching, and also rely on interpolants parameterized via neural networks and adaptations of normalizing flows. 
While closely related to our method, the approach proposed in prior work differs in its formulation for learning the interpolants. Specifically, they employ a specific parameterization for both the interpolant 
$\phi$ and the flow model $v_\theta$, which share  parameters. This design choice, can limit the flexibility of the model. 
Additionally, since they do not use a bi-level objective, the optimization procedure does not account for the relationship between the optimal $\theta$ and optimal $\phi$. In other words, the dependence of $\theta$ on $\phi$ is ignored. Our method does not have these restrictions. Empirically, we observe that our approach achieves improved performance (\cref{sec:exps}), highlighting the benefits of explicitly modeling this relationship. 


\paragraph{Consistency Models}
This family of models \citep{yang2024consistency, yang2023improving, song2023consistency} aims to learn a function that directly solves \cref{ODE-1}. One can choose to solve the ODE numerically and learn the function using a distilled dataset consisting of pairs of noise and the corresponding data-samples, though more common and scalable approaches use a simulation free approach. We solve the problem of learning straight fields, which is orthogonal to the problem that consistency models aim to solve, which involves learning a function that correctly integrates the vector field. In principle, these ideas could be combined to produce more efficient sampling methods.

\section{Experiments} \label{sec:exps}

\begin{figure*}[htbp]
  \centering
  \begin{minipage}[b]{0.3\textwidth}
    \centering
    \includegraphics[width=0.92\textwidth,trim={0.1cm 0 0.1cm 0cm},clip]{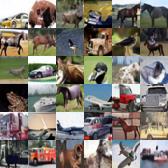} 
    \caption*{CIFAR-10}
    \label{fig:image1}
  \end{minipage}
  \hfill
  \begin{minipage}[b]{0.3\textwidth}
    \centering
    \includegraphics[width=0.95\textwidth]{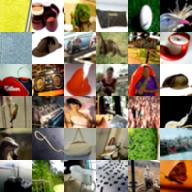} 
    \caption*{ImageNet 32x32}
    \label{fig:image2}
  \end{minipage}
   \hfill
  \begin{minipage}[b]{0.3\textwidth}
    \centering
    \includegraphics[width=0.95\textwidth]{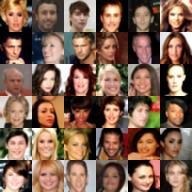} 
    \caption*{CelebA 256}
    \label{fig:image3}
  \end{minipage}
  \caption{Generated samples from CIFAR, ImageNet and CelebA.}
  \label{fig:overall}
\end{figure*}


 We conduct two sets of experiments \footnote{Our code  will be available at \url{https://github.com/sshivs/learn_st_flow/}}: one on low-resolution datasets, including CIFAR-10 \citep{alex2009learning} and ImageNet 32x32 \citep{chrabaszcz2017downsampled}, and another on higher-resolution datasets, namely CelebA-HQ \citep{karras2017progressive} and AFHQ-Cat \citep{choi2020stargan}.  Following the methodology of \citet{song2023consistency}, we evaluate the model across varying numbers of function evaluations (NFE). The flow field is learned using a U-Net architecture based on DDPM++ \citep{song2020score}. To assess the quality of generated images, we employ the Fréchet Inception Distance (FID) score \citep{heusel2017gans}.

\paragraph{Baselines} We follow \citet{song2023consistency,yang2024consistency} and compare our method against several baselines comprising of representative diffusion models and flow models. The baseline models include Consistency Models \citep{song2023consistency}, Rectified Flow \citep{liu2022flow}, Rectified Flow with Bellman Sampling \citep{nguyen2024bellman}, Neural Flow Diffusion Models \citep{bartosh2024neural}, and Consistency-FM \citep{yang2024consistency}. We did not run these baselines ourselves and have reported results from literature. Since not all earlier works have reported results on all the datasets, for each dataset the set of baselines is not always identical. 

\textbf{Results}
The results for the CIFAR dataset are presented in Table \ref{tab:results}. 
Our method demonstrates superior performance compared to models such as Consistency FM \citep{yang2024consistency}, Rectified Flow \citep{liu2022flow}, and Consistency Model \citep{song2023consistency}. We also see that our model matches or outperforms mainstream diffusion models while using a low number of function evaluations (NFE).


\cref{tab:results_imgnet} shows results on the ImageNet 32x32 dataset, where we compare against plain flow matching \citep{lipman2022flow} (with a large number of NFEs), multisample flow matching \citep{pooladian2023multisample}, and Neural Flow Diffusion models \citep{bartosh2024neural}. Our approach yields better performance than competing approaches using a low number of NFEs.


\begin{table}[ht]
    \caption{Comparison with baseline models on CIFAR-10. Results for other models are obtained from previous work.\label{tab:results}}
    \centering
    \begin{tabular}{lcc}
        \toprule
        Method & NFE ($\downarrow$) & FID ($\downarrow$) \\
        \midrule
        Score SDE \cite{song2020score} & 2000 & 2.20 \\
        DDPM \cite{ho2020denoising} & 1000 & 3.17 \\
        LSGM \cite{vahdat2021score} & 147 & 2.10 \\
        PFGM \cite{xu2022poisson} & 110 & 2.35 \\
        EDM \cite{karras2022edm}
         & 35 & \textbf{2.04} \\
         \midrule
        1-Rectified Flow /ReFlow\cite{liu2022flow}
         & 1 & 378 \\
        Glow \cite{kingma2018glow}
         & 1 & 48.9 \\
        Residual Flow \cite{chen2019residual} & 1 & 46.4\\
        GLFlow \cite{xiao2019generative} & 1 & 44.6 \\
        DenseFlow \cite{grcic2021densely} & 1 & 34.9 \\
        Consistency Model  \citep{song2023consistency} & 2 & 5.83 \\
        Consistency Flow Matching \citep{yang2024consistency} & 2 & 5.34 \\
        \midrule
        Ours&2&\textbf{4.61} \\
        \bottomrule
    \end{tabular}
\end{table}

We further evaluate our method on high-resolution image generation tasks, specifically 256$\times$256 images from AFHQ-Cat and CelebA-HQ. Following \citet{yang2024consistency}, we compare against baseline methods, including Consistency FM \citep{yang2024consistency}, ReFlow \citep{liu2022flow}, and ReFlow with Bellman sampling \citep{nguyen2024bellman}. All baseline results are taken from \citet{yang2024consistency}. Our method outperforms baseline approaches such as Rectified Flow \citep{liu2022flow} and Rectified Flow with Bellman sampling \citep{nguyen2024bellman} by a significant margin.

\begin{table}[ht]
	\caption{Comparison with flow matching models on CelebA.\label{tab:results2}}
	\centering
        \begin{tabular}{lcc}
        \toprule
        Method
         & NFE ($\downarrow$) & FID ($\downarrow$) \\
       
        \midrule

        \multirow{ 2}{*}{ReFlow \citep{liu2022flow}}
         &8&109.4\\
        &6&127.0\\
        \midrule
        \multirow{ 2}{*}{ReFlow + Bellman Sampling \citep{nguyen2024bellman}} &8&49.8\\
        &6&72.5\\
        \midrule
        Consistency Flow Matching \citep{yang2024consistency} &6&36.4\\
        \midrule
        Ours &6 & \textbf{28.6}\\
        \bottomrule
	\end{tabular}
\end{table}

\begin{table}[htb!]
	\caption{Comparison with FM models on on ImageNet-32x32.\label{tab:results_imgnet}}
	\centering
        \begin{small}
	\begin{tabular}{lcc}
       \toprule
	    Method & NFE ($\downarrow$) & FID ($\downarrow$) \\
        \midrule
        Flow Matching \citep{lipman2022flow} & 120 & 5.0 \\
        \midrule
        \multirow{ 2}{*}{MultiSample FM \citep{pooladian2023multisample}} & 4 & 17.3 \\
        & 12 & 7.2 \\
        \midrule
        \multirow{ 2}{*}{NFDM \citep{bartosh2024neural}} & 4 & 6.1 \\
        & 12 & 4.1 \\
        \midrule
        \multirow{ 2}{*}{Ours}&4&\textbf{5.58} \\
        &12&\textbf{3.84} \\
        \bottomrule
	\end{tabular}
    \end{small}
\end{table}


\section{Conclusion}
We propose a novel approach to learn flow matching vector fields that; unlike existing methods, which try to learn straight approximations to a curved vector field; learns a straight vector field directly. We use non-linear interpolants in the CFM objective and show how one can optimize the corresponding vector field solutions. In the process, we provide analytical expressions for the general solution to a CFM model and show how it can be tuned to adjust the ``straightness'' of the vector field. We present a way to parametrize the interpolants using a GLOW model, allowing fast inversion and determinant computations. Our approach outperforms recent methods  when using a low number of NFEs.

{
\bibliographystyle{icml25}
\bibliography{mybib,mybib1}
}

\newpage 
\appendix

\end{document}